\documentclass[runningheads, envcountsame, a4paper]{llncs}

\usepackage{pdflscape}
\usepackage[pdftex]{graphicx} 
\usepackage{epstopdf} 
\usepackage[hyphens]{url}
\usepackage{microtype}
\usepackage{amssymb}
\usepackage[utf8]{inputenc}
\usepackage[T1]{fontenc}
\usepackage{amsmath,mathtools}
\usepackage[caption=false]{subfig}

\title{Hybrid Genetic Algorithm and Lasso Test Approach for Inferring Well Supported Phylogenetic Trees based on Subsets of Chloroplastic Core Genes}

\titlerunning{Hybrid Genetic Algorithm and Lasso Test for Inferring Phylogenetic Trees}
\authorrunning{AlKindy B. \emph{et al.}}

\author{Bassam~AlKindy\inst{1,3} \and Christophe~Guyeux\inst{1} \and Jean-Fran\c{c}ois~Couchot\inst{1} \and Michel~Salomon\inst{1} \and Christian Parisod\inst{2} \and Jacques~M.~Bahi\inst{1}}

\institute{$^1$~FEMTO-ST Institute, UMR 6174 CNRS, DISC Computer Science Department, \\ University of Franche-Comt\'{e}, France\\
$^2$~Lab. Evolutionary Botany, University of Neuchâtel, Switzerland\\
$^3$~Department of Computer Science, University of Mustansiriyah, Baghdad, Iraq\\
\email{\{bassam.al-kindy, christophe.guyeux, jean-francois.couchot,	michel.salomon, jacques.bahi\}@univ-fcomte.fr},
\email{christian.parisod@unine.ch}}

\begin{document}
\maketitle
\setcounter{footnote}{0}

\begin{abstract}
The amount of completely sequenced chloroplast genomes increases rapidly every day, leading to the possibility to build large scale phylogenetic trees of plant species. Considering a subset of close plant species defined according to their chloroplasts, the phylogenetic tree that can be inferred by their core genes is not necessarily well supported, due to the possible occurrence of ``problematic'' genes (\textit{i.e.}, homoplasy, incomplete lineage sorting, horizontal gene transfers, \emph{etc.}) which may blur phylogenetic signal. However, a trustworthy phylogenetic tree can still be obtained if the number of problematic genes is low, the problem being to determine the largest subset of core genes that produces the best supported tree. To discard problematic genes and due to the overwhelming number of possible combinations, we propose an hybrid approach that embeds both genetic algorithms and statistical tests. Given a set of organisms, the result is a pipeline of many stages for the production of well supported phylogenetic trees. The proposal has been applied to different cases of plant families, leading to encouraging results for these families. 

\end{abstract}

\section{Introduction}\label{sec:intro}

The multiplication of complete chloroplast genomes should normally lead to the ability to infer trustworthy phylogenetic trees for plant species. Indeed, the existence of trustworthy coding sequence prediction and annotation software specific to chloroplasts (like DOGMA~\cite{RDogma}), together with the good control of sequence alignment and maximum likelihood or Bayesian inference phylogenetic reconstruction techniques, should imply that, given a set of close species, their core genome (the set of genes in common) can be as large and accurately detected as possible to finally obtain a well-supported phylogenetic tree. However, all genes of the core genome are not necessarily constrained in a similar way, some genes having a larger ability to evolve than other ones due to their lower importance. Such minority genes tell their own story instead of the species one, blurring so the phylogenetic information.

To obtain well-supported phylogenetic trees, the deletion of these problematic genes~(which may result from homoplasy, stochastic errors, undetected paralogy, incomplete lineage sorting, horizontal gene transfers, or even hybridization) is needed. A solution is to construct the phylogenetic trees that correspond to all the combinations of core genes, and to finally consider the tree that is as supported as possible while considering as many genes as possible. The major drawback is its inhibitory computational cost, since testing all the possible combinations is totally intractable in practice ($2^n$ phylogenetic tree reconstructions with $n \approx 100$ core genes of plants belonging to the same order). Thus we have to remove the problematic genes without exhaustively testing combinations of genes. Therefore, our proposal is to mix various approaches to extract promising subsets of core genes, encompassing systematic deletion of genes, random selection of large subsets, statistical evaluation of gene effects, and genetic algorithms~(GAs)~\cite{bhandari1996genetic,booker1989classifier}.
These latters are efficient, robust, and adaptive search techniques designed for solving optimization problems, which have the ability to produce semi-optimal solutions~\cite{tate2002parallel,gupta2012novel,matsuda1995construction}.  

The contribution of this article can be summarized as follows. We focus on situations where a large number of genes are shared by a set of species so that, in theory, enough data are available to produce a well supported phylogenetic tree. However, a few genes tell a different evolutionary scenario than the majority of sequences, leading to phylogenetic noise blurring the phylogeny reconstruction. The pipeline that we propose attempts to solve such an issue by computing all phylogenetic trees which can be obtained by removing at most one core gene. In case where such a preliminary systematic approach does not solve the phylogeny, new investigation stages are added to the pipeline, namely a Monte-Carlo based random approach and two invocations of a genetic algorithm, separated by a Lasso test. The pipeline is finally tested on various sets of chloroplast genomes. 

The remainder of this article is as follows. We start with Section~\ref{section2} by giving a brief and global description of the problem. Genetic population initialization is discussed in Section~\ref{section3}, while the first optimization stage with genetic algorithm is fully detailed in Section~\ref{section4}. Targeting problematic genes using a Lasso test and the following second invocation of the genetic algorithm is detailed in Section~\ref{removegenes}. Then, in the next section, various plant families are tested as a case study. Finally this research work ends with a conclusion section in which the contributions are summarized and intended future work is outlined.

\section{Presentation of the problem}\label{section2}

Let us consider a set of chloroplast genomes that have been annotated using DOGMA~\cite{RDogma} (\url{http://dogma.ccbb.utexas.edu/}). We have then access to the core genome~\cite{Alkindy2014} (genes present everywhere) of these species, whose size is about one hundred genes when the species are close enough. For further information on how we found the core genome, see~\cite{Alkindy2014,Alkindy_BIBM2014}.
Sequences are further aligned with  MUSCLE~\cite{edgar2004muscle} and the RAxML~\cite{stamatakis2005raxml} tool infers the corresponding phylogenetic tree. If this resulting tree is well-supported, then the process is stopped without further investigations. Indeed,  
if all bootstrap values are larger than 95, then we can reasonably consider that the phylogeny of these species is resolved, as the largest possible number of genes has led to a very well supported tree.

\begin{figure}[!ht]
\begin{center}
    \includegraphics[width=0.8\textwidth]{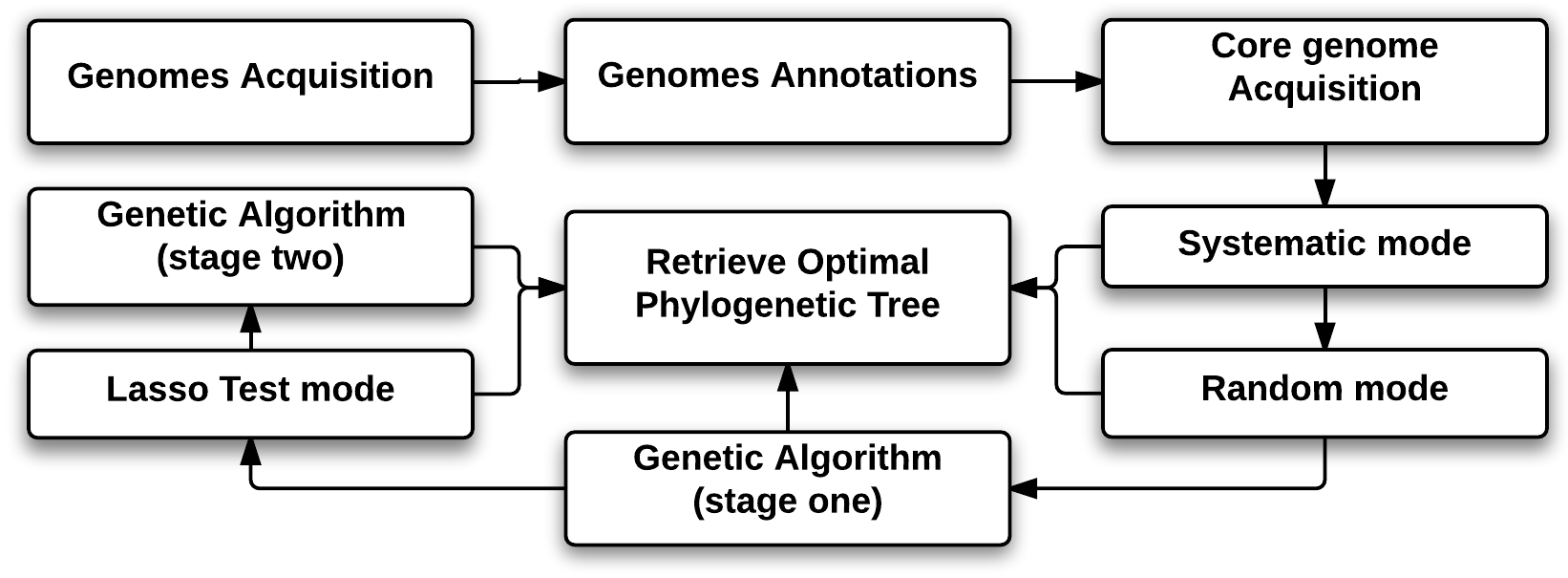}
    \caption{Overview of the proposed pipeline for phylogenies based on chloroplasts.}\label{fig:GA_generalview}
\end{center}
\end{figure}

In case where some branches are not supported well, we can wonder whether a few genes can be incriminated in this lack of support. If so, we face an optimization problem: \emph{find the most supported tree using the largest subset of core genes}. Obviously, a brute force approach investigating all possible combinations of genes is intractable, as it leads to $2^n$ phylogenetic tree inferences for a core genome of size $n$. To solve this optimization problem, we have proposed an hybrid approach mixing a genetic algorithm with the use of some statistical tests for discovering problematic genes. The initial population for the genetic algorithm is built by both systematic and random pre-GA investigations. These considerations led to a pipeline detailed in Figure~\ref{fig:GA_generalview}, whose stages will be developed thereafter.

\begin{figure}[!ht]
\begin{center}
    \subfloat[Systematic mapping.\label{fig:GA_map_sysematic}]{%
    \includegraphics[width=0.4\textwidth]{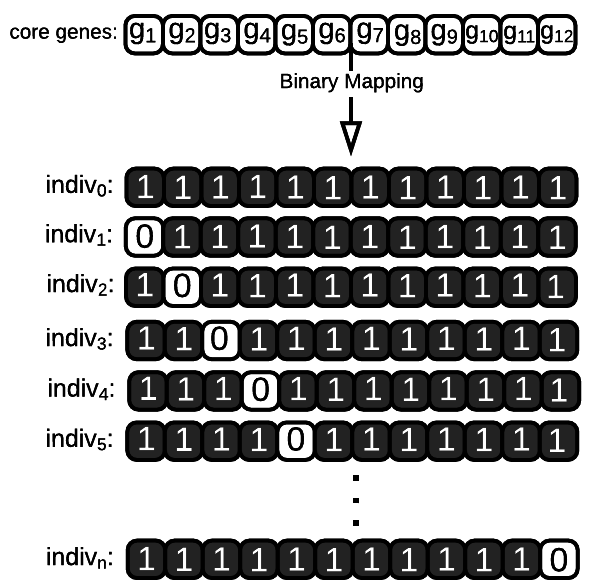}
    }
    \subfloat[Random mapping.\label{fig:GA_map_random}]{%
      \includegraphics[width=0.4\textwidth]{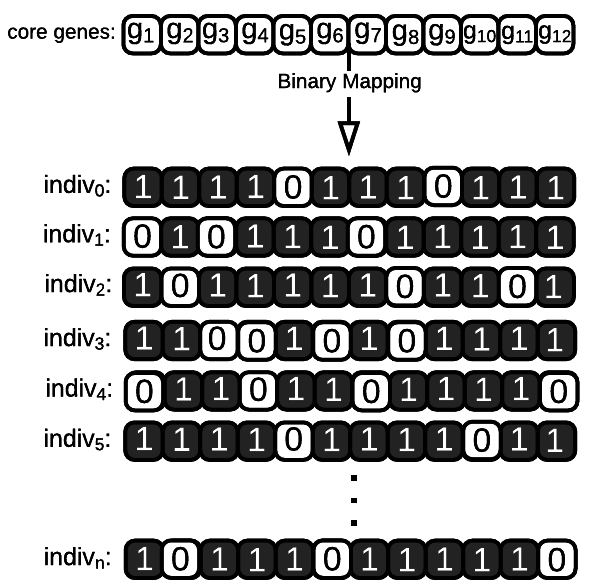} 
    }
  \caption{Binary mapping operation and genetic algorithm overview. (a) Initial individuals obtained in systematic mode stage. Two kinds of individuals are generated. First, by considering all genes in the core genome. Second, by omitting one gene sequentially depending on the core length.  (b) Initial individuals are generated randomly in random stage (random mode) by omitting 2-10 genes randomly.}
    \label{fig:GA_binaryMap}
\end{center}
\end{figure}

\section{Generation of the initial population}\label{section3}

The objective is to obtain a well-supported phylogenetic tree by using the largest possible subset of genes. If this goal cannot be reached by taking all core genes, the first thing to investigate is to check whether one particular gene is responsible of this problem. Therefore we systematically compute all the trees we can obtain by removing exactly one gene from the core genome, leading to $n$ new phylogenetic trees, where $n$ is the core size (see Figure~\ref{fig:GA_map_sysematic}).

If, during this systematic approach, one well-supported tree is obtained, then it is returned as the phylogeny of the species under consideration. Conversely, if all trees obtained have at least one problematic branch, then deeper investigations are required. However the systematic approach has reached its limits which is preliminary to GA, as investigating all phylogenetic trees that can be obtained by removing randomly 2 genes among a core genome of size $n$ leads to $\frac{n(n-1)}{2}$ tree inferences. Obviously, the number of cases explodes, and it is illusory to hope to investigate all reachable trees by discarding 10\% of a core genome having 100 genes. This is why a genetic algorithm has been proposed.

Using the $n+1$ computed trees to initialize the population of the genetic algorithm results in a population which remains too small and too homogeneous. Indeed, all these trees have been computed in the same way, each inference being produced using 99\% of the core genome (we have removed at most 1 gene in a core genome having approximately 100 genes). Thus, in order to increase the diversity of the initial population a second stage (random stage) as shown in Figure~\ref{fig:GA_map_random}, which extracts large random subsets of core genome for inference, is applied.

More precisely, there are indeed two random stages. The first one operates during 200 iterations: at each iteration, an integer $k$ between 2 and 10 is first randomly picked. Then $k$ genes are randomly removed from the core genome, and a phylogeny is inferred using the remaining genes. If during these iterations, by chance, a very well supported tree is obtained, a stop signal is sent to the master process and the obtained tree is returned. If not, we now have enough data to build a relevant initial population for the genetic algorithm. And the second random stage is indeed included in this genetic algorithm.

\section{Genetic algorithm}\label{section4}

A genetic algorithm is a well-known metaheuristic which has been described by a rich body of literature since its introduction in the mid-seventies~\cite{Holland:1975,Holland:1992:ANA:129194}. In the following, we will only discuss the choices we made regarding operators and parameters. For further information and applications regarding the genetic algorithm, see, \emph{e.g.},~\cite{bhandari1996genetic,booker1989classifier,prebys2007genetic,goldberg1993}.

\subsection{Genotype and fitness value}

Genes of the core genome are supposed to be ordered lexicographically. At each subset $s$ of the core genome corresponds thus a unique binary word $w$ of length $n$: for each $i$ lower than $n$, $w_i$ is 1 if the $i$-th core gene is in $s$, else $w_i$ is equal to 0. At each binary word $w$ of length $n$, we can associate its percentage $p$ of 1's and the lowest bootstrap $b$ of the phylogenetic tree we obtain when considering the subset of genes associated to $w$. At each word $w$ we can thus associate as fitness value the score $b+p$, which must be as large as possible. 
We currently consider that bootstrap $b$ and the number of genes $p$ have the same importance in the scoring function. However, changing the weight of each parameter may be interesting in deeper investigations.

\subsection{Genetic process}

\begin{figure}[!ht]
\begin{center}
    \includegraphics[width=0.6\textwidth]{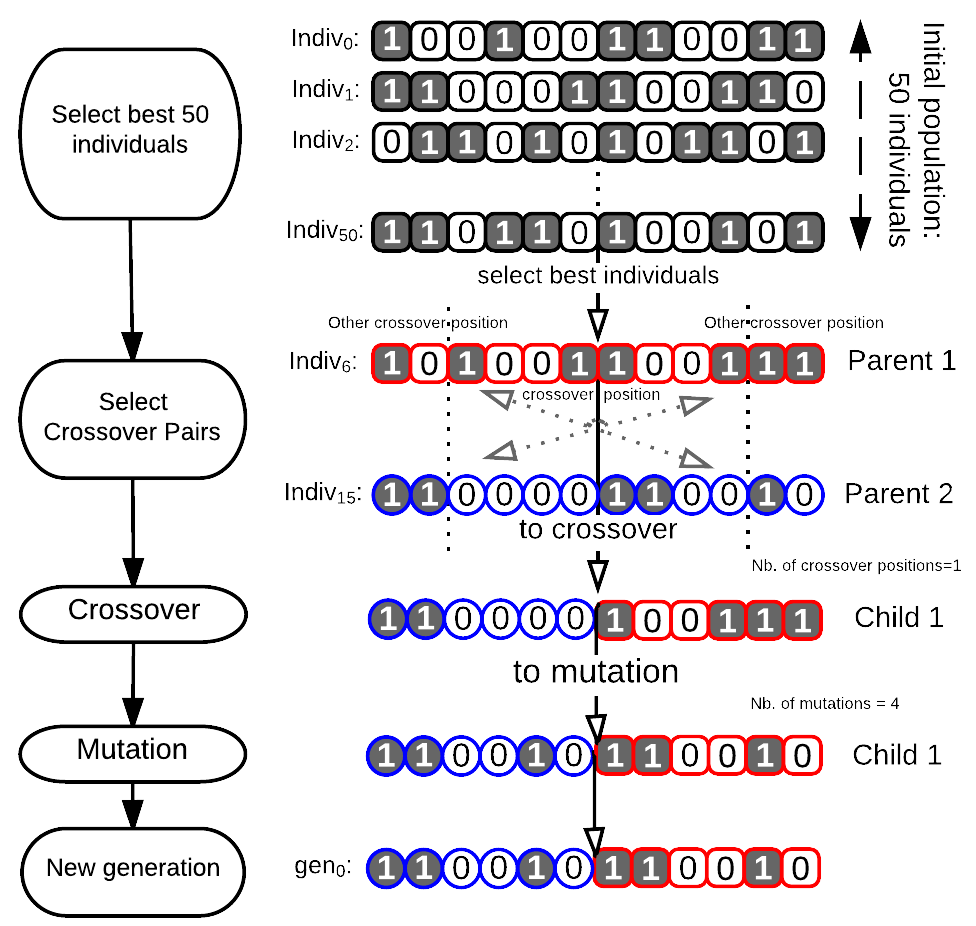}
    \caption{Outlines of genetic algorithm.}\label{fig:GA_operations}
\end{center}
\end{figure}

Until now, binary words (genotypes) of length $n$ that have been investigated are:
\begin{enumerate}
\item the word having only 1's (systematic mode);
\item all words having exactly one 0 (systematic mode);
\item 200 words having between 2 and 10 0's randomly located (random mode).
\end{enumerate}
To each of these words is attached a score which is used to select the 50 best words, or fittest individuals, in order to build the initial population. After that, the genetic algorithm will loop during 200 iterations or until an offspring word such that $b\geqslant 95$ is obtained. During an iteration the algorithm will apply the following steps to produce a new population $P'$  given a population $P$ (see Figure~\ref{fig:GA_operations}).

\begin{itemize}
\item Repeat 5 times a random pickup of a couple of words and mix them using a crossover approach. The obtained words are added to the population $P$, as described in Section~\ref{sec:crossover}, resulting in population $P_c$.
\item Mutate 5 words of the population $P_c$, the mutated words being added too to $P_c$, as detailed in Section~\ref{sec:mut}, leading to population $P_m$.
\item Add 5 new random binary words having less than 10\% of 0's (see Section~\ref{sec:random}) to $P_m$ producing population $P_r$.
\item Select the 50 best words in population $P_r$ to form the new population~$P'$.
\end{itemize}
Let us now explain with more details each step of this genetic algorithm.

\subsection{Crossover step} \label{sec:crossover}

\begin{figure}[!ht]
\begin{center}
    \subfloat[Crossover operation.\label{subfig:crossover}]{%
    \includegraphics[width=0.475\textwidth]{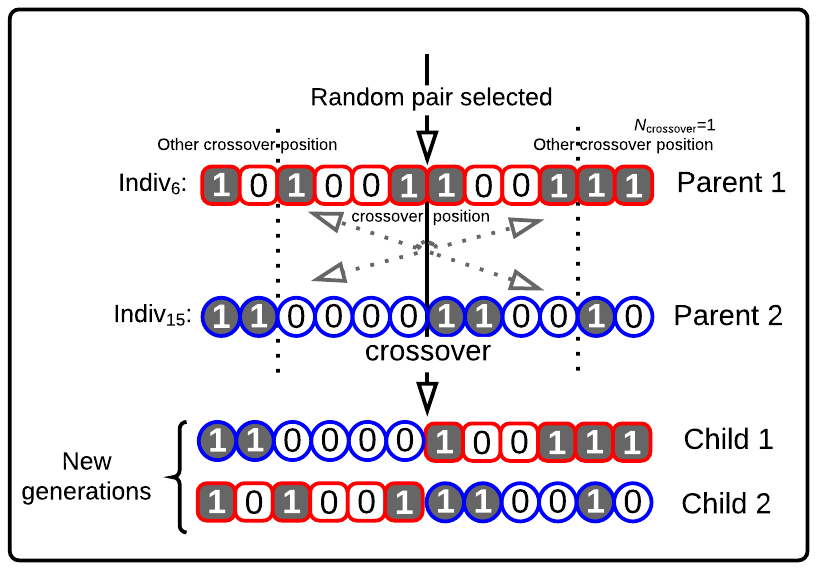} 
    }
    \hspace{0.5em}
    \subfloat[Mutation operation.\label{subfig:mutation}]{%
      \includegraphics[width=0.475\textwidth]{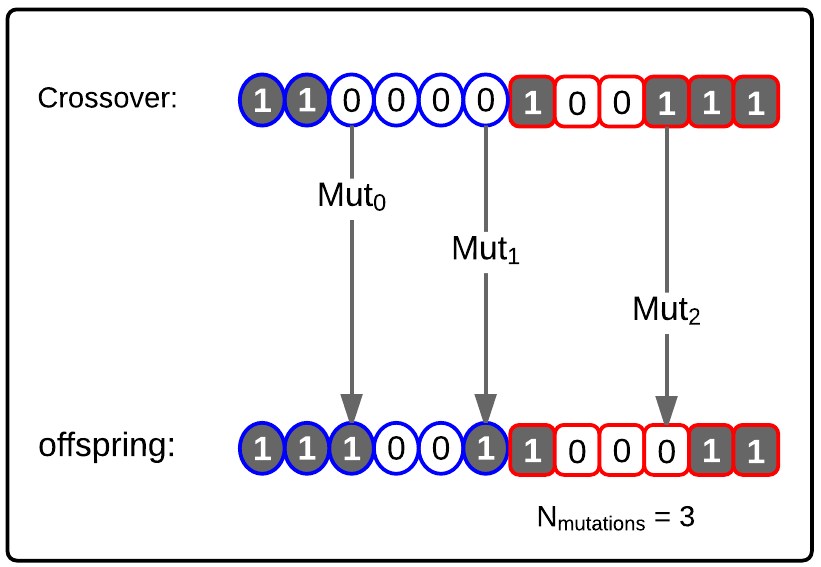} 
    }
    \caption{(a) Two individuals are selected from given population. First portion from determined crossover position in the first individual is switched with the first portion of the second individual. The number of crossover positions is determined by $N_{crossover}$. (b) Random mutations are applied depending on the value of $N_{mutation}$, changing randomly gene state from 1 to 0 or vice versa. New offsprings generated from this stage are predicted w.r.t natural evolution scenario.}
    \label{fig:GA_op}
    \end{center}
\end{figure}

Given two words $w^1$ and $w^2$, the idea of the crossover operation is to mix them, hoping by doing so to generate a new word $w$ having a better score (see Figure~\ref{subfig:crossover}). For instance, if we consider a one-point crossover located at the middle of the words, for $i<\frac{n}{2}$, $w_i=w_i^1$, while for $i\geqslant \frac{n}{2}$, $w_i=w_i^2$: in that case, for the first core genes, the choice (to take them or not for phylogenetic construction) in $w$ is the same than in $w^1$, while the subset of considered genes in $w$ corresponds to the one of $w^2$ for the last 50\% of core genes.

More precisely, at each crossover step, we first pick randomly an integer $k$ lower than $\frac{n}{2}$, and randomly again $k$ different integers $i_1, \hdots i_k$ such that $1<i_1 < i_2 < \hdots < i_k < n$. Then $w^1$ and $w^2$ are randomly selected from the population~$P$, and a new word $w$ is computed as follows:
\begin{itemize}
\item $w_i = w_i^1$ for $i=1,...,i_1$,
\item $w_i = w_i^2$ for $i=i_1+1,...,i_2$,
\item $w_i = w_i^1$ for $i=i_2+1,...,i_3$,
\item etc.
\end{itemize}
Then the phylogenetic tree based on the subset of core genes labeled by $w$ is computed, the score $s$ of $w$ is deduced, and $w$ is added to the population with the fitness value $s$ attached to it.

\begin{figure}[!ht]
\begin{center}
    \includegraphics[width=0.9\textwidth]{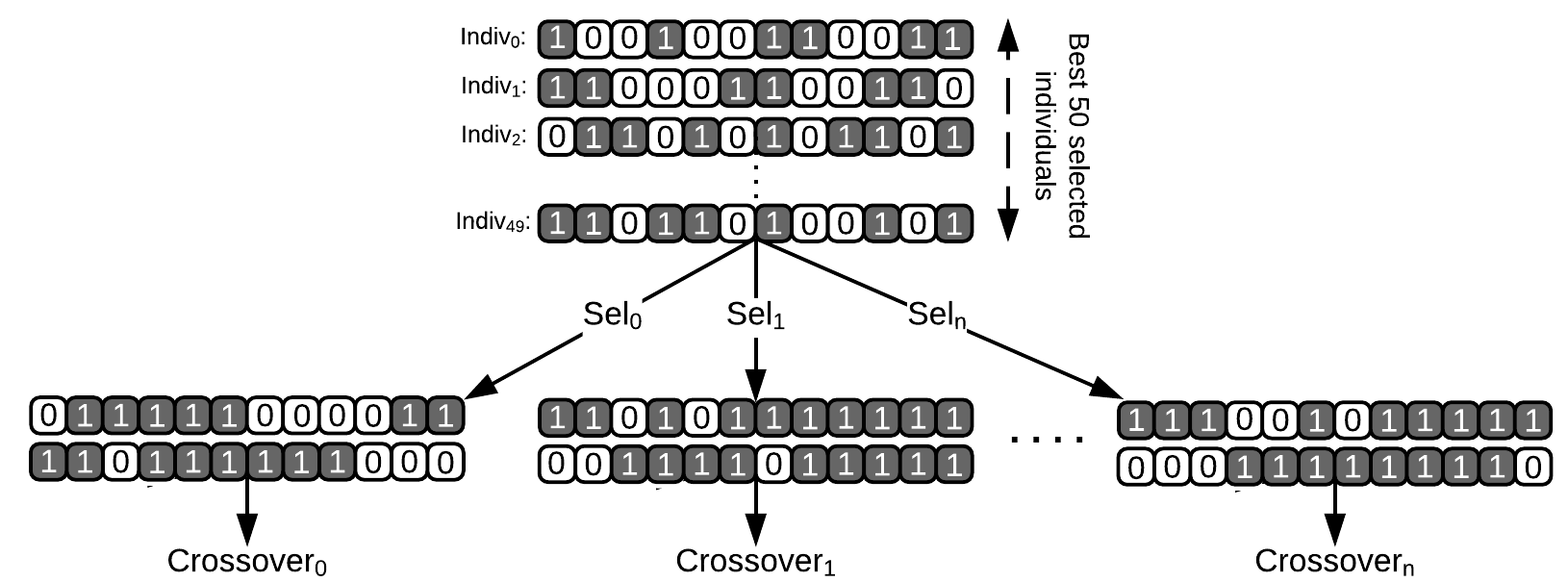}
    \caption{Random pair selections from given population.}\label{fig:GA_selection}
\end{center}
\end{figure}


\subsection{Mutation step}\label{sec:mut}

In this step, we ask whether changing a little a given subset of genes, by removing a few genes and adding a few other ones, may by chance improve the support of the associated tree. Similarly speaking, we try here to improve the score of a given word by replacing a few 0's by 1 and a few 1's by 0. 

In practice, an integer $k \leqslant \frac{n}{4}$ corresponding to the number of changes, or ``mutations'', is randomly picked. Then $k$ different integers $i_1, \hdots, i_k$ lower than $n$ are randomly chosen and a word $w$ is randomly extracted from the current population. A new word $w'$ is then constructed as follows: for each $i=1,...,n$,
\begin{itemize}
\item if $i$ in  $\{i_1, \hdots, i_k\}$, then $w'_i = w_i+1 \mod 2$ (the bit is mutated),
\item else  $w'_i = w_i$ (no modification).
\end{itemize}

\noindent Again, the phylogenetic tree corresponding to the subset of core genes associated to $w'$ is computed, and $w'$ is added to the population together with its score.

\subsection{Random step} \label{sec:random}

In this step, new words having a large amount of 1's are added to the population. Each new word is obtained by starting from the word having $n$ 1s, followed by $k$ random selection of 1s which are changed to 0, where $k$ is an integer randomly chosen between 1 and 10. The new word is added to the population after having computed its score thanks to a phylogenetic tree inference.

\section{Targeting problematic genes using statistical tests}\label{removegenes}

\subsection{The Lasso test}

After having carried out 200 iterations of the genetic algorithm detailed above, it may occur that no well-supported tree has been produced. Various reasons may explain this failure, like a lazy convergence speed, a large number of problematic genes (\emph{e.g.}, homoplasic ones, or due to stochastic errors, undetected paralogy, incomplete lineage sorting, horizontal gene transfers, or hybridization), or close divergence species leading to very small branch lengths between two internal nodes. However, we now have computed enough word scores to determine the effects of each gene in topologies and bootstraps, and to remove the few genes that break supports.

The idea is then to investigate each topology that has appeared enough times during previous computations. In this study, we only consider topologies having a frequency of occurrence larger than 10\%. Remark that this 10\% is convenient for the given case study, but it must depend in fact on the number of obtained topologies. Then for each best word of these best topologies, and for each problematic bootstrap in its associated tree, we apply a Lasso approach as follows.

The Lasso~(Least Absolute Shrinkage and Selection Operator) test~\cite{tibshirani96regression} is an estimator that takes place in the category of least-squares regression analysis. Like all the algorithms in this group, it estimates a linear model which minimizes a residual sum with respect to a variable $\lambda$.
Let us explain how this variable can be used to order genes with respect to their ability to modify the bootstrap support.

Let $X$ be a $m\times p$ matrix where each line $X_i =(X_{i1}, \ldots ,X_{ij},\ldots X_{ip})$, $1 \le i \le m$, is a configuration where 
$X_{ij}$ is  1 if gene number $j$ is present inside the configuration $i$ and $X_{ij}$ is  0 otherwise.
For each $X_i$, let $Y_i$ be the real positive support value for each problematic bootstrap $b$ per topology and per gene. According to~\cite{tibshirani96regression}, the Lasso test $\beta=(\beta_1, \ldots ,\beta_i,\ldots \beta_{p})$ is defined by 

\newcommand{\argmin}{\operatornamewithlimits{argmin}}

\begin{equation}
\beta = \argmin \left\{
\sum_{i=1}^m \left( Y_i - \sum_{j=1}^p \beta_j X_{ij} \right)^2 + \lambda\sum_{j=1}^{p} | \beta_j |  
\right\} .
\end{equation}

When $\lambda$ has high value, all the $\beta_j$ are null.
It is thus sufficient to decrease the value of $\lambda$ to observe that some $\beta_j$ become not null.
Moreover, the sign of $\beta_j$ is positive (resp. negative) if the bootstrap support increases (resp. decreases) with respect to $j$.

\subsection{Second stage of genetic algorithm}\label{final}

Targeting problematic genes using Lasso approach can solve the issue of badly supported values in some cases, especially when only one support is lower than the predefined threshold. In cases where at least two branches are not well supported, removing genes that break the first support may or may not has an effect on the second problematic support. In other words, each of the two problematic supports can be separately solved using Lasso investigations, but not necessarily both together.

However, the population has been improved, receiving very interesting words for each problematic branch. Then a last genetic algorithm phase is launched on the updated population, in order to mix these promising words by crossover operations, hoping by doing so to solve in parallel all of the badly supported values. This last stage runs until either the resolution of all problematic bootstraps or the reach of iterations limit (set to 1000 in our simulations).

\section{Case studies}\label{case}

\subsection{Pipeline evaluation on various groups of plant species}

In this section, the proposed pipeline is tested on various sets of close plant species. An example of 50 subgroups (ranging on average from 12 to 15 chloroplasts species) encompassing 356 plant species is presented in Table~\ref{families}. The \emph{Stage} column contains the termination step for each subgroup, namely: the systematic~(code 1), random~(2), or optimization stages~(3) using genetic algorithm and/or Lasso test. A large occurrence in this table means that the associated group and/or subgroups has its computation terminated in either penultimate or last pipeline stage. An occurrence of 31 is frequent due to the fact that 32 MPI threads (one master plus 31 slaves) have been launched on our supercomputer facilities. Notice that the Table~\ref{families} is divided 
into four parts: 
groups of species stopped in systematic stage with weak bootstrap values (which is due to the fact that a upper time limit has been set for each group and/or subgroups, while each computed tree in these remarkable groups needed a lot of times for computations), 
subgroups terminated during systematic stage with desired bootstrap value, 
groups or subgroups terminated in random stage with desired bootstrap value, and finally, 
groups or subgroups terminated during  optimization stages. 
The majority of subgroups has its phylogeny satisfactorily resolved, as can be seen on all obtained trees which are downloadable at \url{http://meso.univ-fcomte.fr/peg/phylo}. 
In what follows, an example of problematic group, namely the \textit{Apiales}, is more deeply investigated as a case study. 


\begin{table}[!ht]
\caption{Families applied on pipeline stages}
\begin{center}
\scalebox{0.70}{%
\begin{tabular}{l||c|c|c|c|c|c|l}
\multicolumn{1}{c||}{Group or subgroup} & Occurrences & Core genes & \# Species & L.Bootstrap & Pip. Stage & Likelihood & \multicolumn{1}{c}{Outgroup} \\ \hline\hline
\textit{Gossypium\_group\_0} & 85 & 84 & 12 & 26 & 1 & -84187.03 & \textit{Theo\_cacao} \\ 
\textit{Ericales} & 674 & 84 & 9 & 67 & 3 & -86819.86 & \textit{Dauc\_carota} \\ 
\textit{Eucalyptus\_group\_1} & 83 & 82 & 12 & 48 & 1 & -62898.18 & \textit{Cory\_gummifera} \\ 
\textit{Caryophyllales} & 75 & 74 & 10 & 52 & 1 & -145296.95 & \textit{Goss\_capitis-viridis} \\ 
\textit{Brassicaceae\_group\_0} & 78 & 77 & 13 & 64 & 1 & -101056.76 & \textit{Cari\_papaya} \\ 
\textit{Orobanchaceae} & 26 & 25 & 7 & 69 & 1 & -19365.69 & \textit{Olea\_maroccana} \\ 
\textit{Eucalyptus\_group\_2} & 87 & 86 & 11 & 71 & 1 & -72840.23 & \textit{Stoc\_quadrifida} \\ 
\textit{Malpighiales} & 1183 & 78 & 12 & 80 & 3 & -95077.52 & \textit{Mill\_pinnata} \\ 
\textit{Pinaceae\_group\_0} & 76 & 75 & 6 & 80 & 1 & -76813.22 & \textit{Juni\_virginiana} \\ 
\textit{Pinus} & 80 & 79 & 11 & 80 & 1 & -69688.94 & \textit{Pice\_sitchensis} \\ 
\textit{Bambusoideae} & 83 & 81 & 11 & 80 & 3 & -60431.89 & \textit{Oryz\_nivara} \\ 
\textit{Chlorophyta\_group\_0} & 231 & 24 & 8 & 81 & 3 & -22983.83 & \textit{Olea\_europaea} \\ 
\textit{Marchantiophyta} & 65 & 64 & 5 & 82 & 1 & -117881.12 & \textit{Pice\_abies} \\ 
\textit{Lamiales\_group\_0} & 78 & 77 & 8 & 83 & 1 & -109528.47 & \textit{Caps\_annuum} \\ 
\textit{Rosales} & 81 & 80 & 10 & 88 & 1 & -108449.4 & \textit{Glyc\_soja} \\ 
\textit{Eucalyptus\_group\_0} & 2254 & 85 & 11 & 90 & 3 & -57607.06 & \textit{Allo\_ternata} \\ 
\textit{Prasinophyceae} & 39 & 43 & 4 & 97 & 1 & -66458.26 & \textit{Oltm\_viridis} \\ 
\textit{Asparagales} & 32 & 73 & 11 & 98 & 1 & -88067.37 & \textit{Acor\_americanus} \\ 
\textit{Magnoliidae\_group\_0} & 326 & 79 & 4 & 98 & 3 & -85319.31 & \textit{Sacc\_SP80-3280} \\ 
\textit{Gossypium\_group\_1} & 66 & 83 & 11 & 98 & 1 & -81027.85 & \textit{Theo\_cacao} \\ 
\textit{Triticeae} & 40 & 80 & 10 & 98 & 1 & -72822.71 & \textit{Loli\_perenne} \\ 
\textit{Corymbia} & 90 & 85 & 5 & 98 & 2 & -65712.51 & \textit{Euca\_salmonophloia} \\ 
\textit{Moniliformopses} & 60 & 59 & 13 & 100 & 1 & -187044.23 & \textit{Prax\_clematidea} \\ 
\textit{Magnoliophyta\_group\_0} & 31 & 81 & 7 & 100 & 1 & -136306.99 & \textit{Taxu\_mairei} \\ 
\textit{Liliopsida\_group\_0} & 31 & 73 & 7 & 100 & 1 & -119953.04 & \textit{Drim\_granadensis} \\ 
\textit{basal\_Magnoliophyta} & 31 & 83 & 5 & 100 & 1 & -117094.87 & \textit{Ascl\_nivea} \\ 
\textit{Araucariales} & 31 & 89 & 5 & 100 & 1 & -112285.58 & \textit{Taxu\_mairei} \\ 
\textit{Araceae} & 31 & 75 & 6 & 100 & 1 & -110245.74 & \textit{Arun\_gigantea} \\ 
\textit{Embryophyta\_group\_0} & 31 & 77 & 4 & 100 & 1 & -106803.89 & \textit{Stau\_punctulatum} \\ 
\textit{Cupressales} & 87 & 78 & 11 & 100 & 2 & -101871.03 & \textit{Podo\_totara} \\ 
\textit{Ranunculales} & 31 & 71 & 5 & 100 & 1 & -100882.34 & \textit{Cruc\_wallichii} \\ 
\textit{Saxifragales} & 31 & 84 & 4 & 100 & 1 & -100376.12 & \textit{Aral\_undulata} \\ 
\textit{Spermatophyta\_group\_0} & 31 & 79 & 4 & 100 & 1 & -94718.95 & \textit{Mars\_crenata} \\ 
\textit{Proteales} & 31 & 85 & 4 & 100 & 1 & -92357.77 & \textit{Trig\_doichangensis} \\ 
\textit{Poaceae\_group\_0} & 31 & 74 & 5 & 100 & 1 & -89665.65 & \textit{Typh\_latifolia} \\ 
\textit{Oleaceae} & 36 & 82 & 6 & 100 & 1 & -84357.82 & \textit{Boea\_hygrometrica} \\ 
\textit{Arecaceae} & 31 & 79 & 4 & 100 & 1 & -81649.52 & \textit{Aegi\_geniculata} \\ 
\textit{PACMAD\_clade} & 31 & 79 & 9 & 100 & 1 & -80549.79 & \textit{Bamb\_emeiensis} \\ 
\textit{eudicotyledons\_group\_0} & 31 & 73 & 4 & 100 & 1 & -80237.7 & \textit{Eryc\_pusilla} \\ 
\textit{Poeae} & 31 & 80 & 4 & 100 & 1 & -78164.34 & \textit{Trit\_aestivum} \\ 
\textit{Trebouxiophyceae} & 31 & 41 & 7 & 100 & 1 & -77826.4 & \textit{Ostr\_tauri} \\ 
\textit{Myrtaceae\_group\_0} & 31 & 80 & 5 & 100 & 1 & -76080.59 & \textit{Oeno\_glazioviana} \\ 
\textit{Onagraceae} & 31 & 81 & 5 & 100 & 1 & -75131.08 & \textit{Euca\_cloeziana} \\ 
\textit{Geraniales} & 31 & 33 & 6 & 100 & 1 & -73472.77 & \textit{Ango\_floribunda} \\ 
\textit{Ehrhartoideae} & 31 & 81 & 5 & 100 & 1 & -72192.88 & \textit{Phyl\_henonis} \\ 
\textit{Picea} & 31 & 85 & 4 & 100 & 1 & -68947.4 & \textit{Pinu\_massoniana} \\ 
\textit{Streptophyta\_group\_0} & 31 & 35 & 7 & 100 & 1 & -68373.57 & \textit{Oedo\_cardiacum} \\ 
\textit{Gnetidae} & 31 & 53 & 5 & 100 & 1 & -61403.83 & \textit{Cusc\_exaltata} \\ 
\textit{Euglenozoa} & 29 & 26 & 4 & 100 & 3 & -8889.56 & \textit{Lath\_sativus} \\ \hline\hline
\end{tabular}}
\end{center}
\label{families}
\end{table}

\subsection{Investigating \textit{Apiales} order}

In our study \textit{Apiales} choroplasts consist of two sets, as detailed in Table~\ref{table:population}: two species belong to the \textit{Apiaceae} family set (namely \textit{Daucus carota} and \textit{Anthriscus cerefolium}), while the remaining seven species are in the \textit{Araliaceae} family set. These latter are: \textit{Panax ginseng, Eleutherococcus senticosus, Aralia undulata, Brassaiopsis hainla, Metapanax delavayi, Schefflera delavayi}, and \textit{Kalopanax septemlobus}. Chloroplasts of \textit{Apiales} are characterized by having highly conserved gene content and order~\cite{palmer1991plastid}. 

\begin{table}[!ht]
\caption{Genomes information of Apiales}
\begin{center}
\scalebox{0.85}{
\begin{tabular}{l|c|c|c|c|c}
\multicolumn{1}{c|}{Organism name} & Accession & Genome Id & Sequence length & Number of genes & Lineage \\ \hline\hline
\textit{Daucus carota} & NC\_008325.1 & 114107112 & 155911 bp & 138 & Apiaceae \\ 
\textit{Anthriscus cerefolium} & NC\_015113.1 & 323149061 & 154719 bp & 132 & Apiaceae \\ 
\textit{Panax ginseng} & NC\_006290.1 & 52220789 & 156318 bp & 132 & Araliaceae \\ 
\textit{Eleutherococcus senticosus} & NC\_016430.1 & 359422122 & 156768 bp & 134 & Araliaceae \\ 
\textit{Aralia undulata} & NC\_022810.1 & 563940258 & 156333 bp & 135 & Araliaceae \\ 
\textit{Brassaiopsis hainla} & NC\_022811.1 & 558602891 & 156459 bp & 134 & Araliaceae \\ 
\textit{Metapanax delavayi} & NC\_022812.1 & 558602979 & 156343 bp & 134 & Araliaceae \\ 
\textit{Schefflera delavayi} & NC\_022813.1 & 558603067 & 156341 bp & 134 & Araliaceae \\ 
\textit{Kalopanax septemlobus} & NC\_022814.1 & 563940364 & 156413 bp & 134 & Araliaceae \\ \hline\hline
\end{tabular}}
\end{center}
\label{table:population}
\end{table}

\begin{figure}[!hb]
\centering
    \includegraphics[width=1\textwidth]{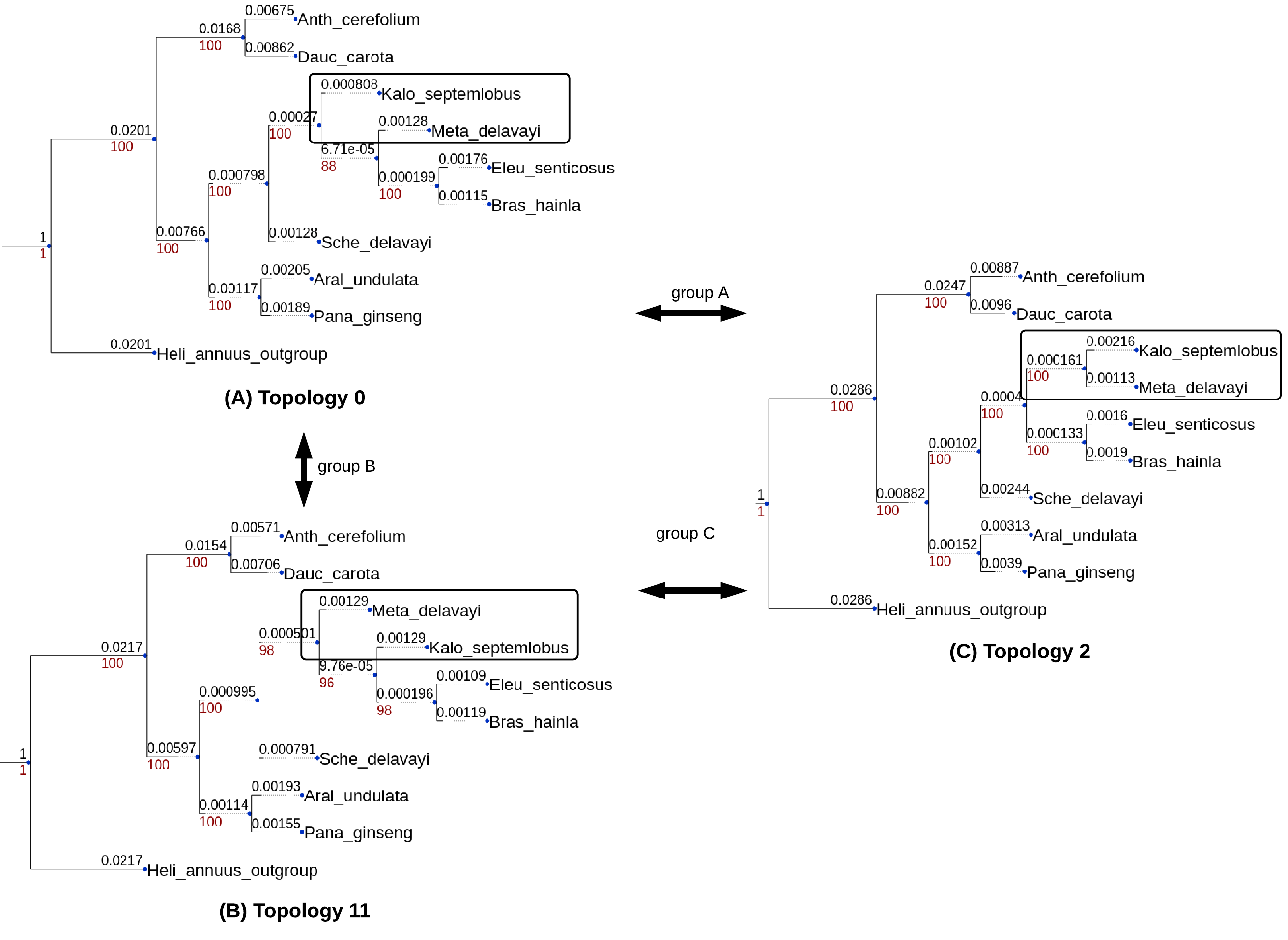}
    \caption{Best trees of topologies 0, 11, and 2.}
    \label{fig:topologies}
\end{figure}

\subsubsection{Method to select best topologies}

We define $T=[t_{0}, t_{1},..., t_{m}]$ as a list of $m=9,053$ obtained trees from given pipeline. By comparing each tree $t_i$ in $T$ with the other trees in $T$, a set of topologies is then numbered and defined as $W=\{w_{0}, w_{1}, w_{2},...,w_{n}\}$, where $w_{i}$ is the topology of number $i$. Let $f(x)$ be a function on W which represents the number of trees having $x$ for their topology. We say that a given topology $w_{i}$ is selected as the best topology if and only if $f(w_{i})\geq lb$ where $lb$ is the lower bound threshold computed by the following formula $$lb=\frac{m*\gamma}{100}$$ $\gamma$ is a constant value between $1-10$ and $m$ is the size of $T$. Then $x$ is stored as best topology. 

\subsubsection{Pratical results}

In our case, $\gamma=8$, meaning that we exclude as noise the topologies representing less than 8\% from the given trees. Three from 43 identified tree topologies are selected, with a number of occurrences $f(x)$ above $lb=724$, as the best topologies as shown in Table~\ref{Topologies}. In this table, topologies~0 and 11 are delivered from optimization stages when the desired bootstrap value is set to 96, and topology~2 is obtained from systematic stage when we increase the desired bootstrap to 100. 
The best obtained phylogenetic trees from selected topologies are provided in Table~\ref{Topologies}: in this table
\emph{Min.Bootstrap} is higher than \emph{Avg.Bootstrap}, as the former represents the lowest bootstrap value 
of the best tree in the given topology, while \emph{Avg.Bootstrap} consists of the average lowest bootstrap in all trees having this topology.

\begin{table}[!ht]
\caption{Information regarding obtained topologies}
\begin{center}
\begin{tabular}{c|c|c|c|c}
Topology & Min.Bootstrap & Avg.Bootstrap & Occurrences ($f(x)$) & Gene rate~(\%)\\ 
\hline\hline
0 & 88 & 56 & 5422 & 64.7\\ 
11 & 96 & 76 & 2579 & 44.8\\ 
2 & 100 & 68 & 787 & 99.1\\ 
8 & 72 & 50 & 89 & 44.8\\ 
9 & 49 & 29 & 48 & 35.3\\ 
14 & 61 & 48 & 31 & 25\\ 
5 & 80 & 48 & 21 & 34.5\\ 
20 & 63 & 53 & 11 & 53.4\\ 
10 & 62 & 50 & 8 & 68.1\\ 
\hline\hline
\end{tabular}
\end{center}
\label{Topologies}
\end{table}

\begin{figure}[!ht]
\centering
    \subfloat[ \label{subfig:tpt}]{%
    \includegraphics[width=0.5\textwidth]{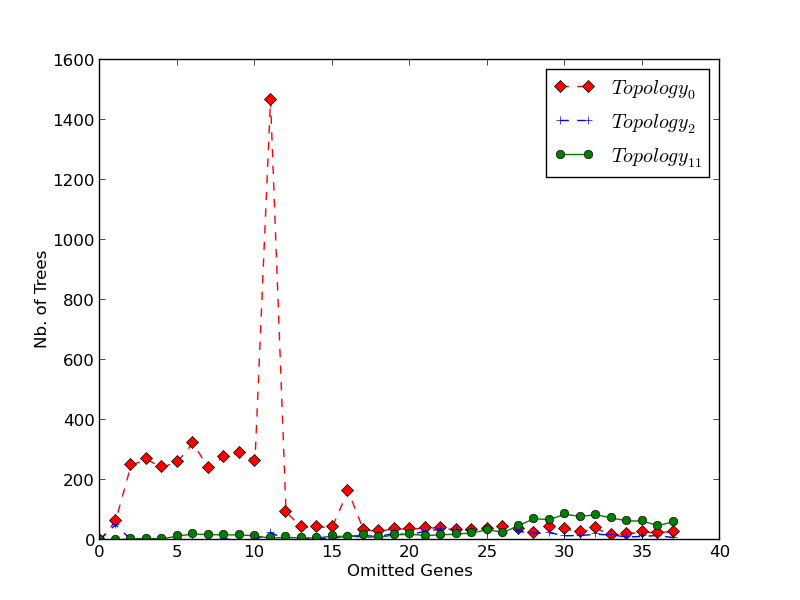}
    }
    \subfloat[ \label{subfig:lbslt}]{%
      \includegraphics[width=0.5\textwidth]{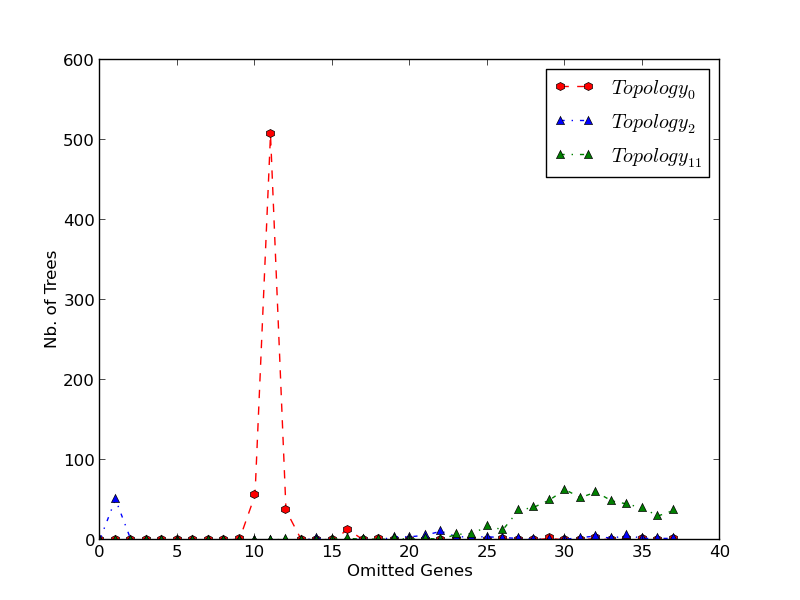}
    }
    \hfill
    \subfloat[\label{subfig:lbs}]{%
      \includegraphics[width=0.5\textwidth]{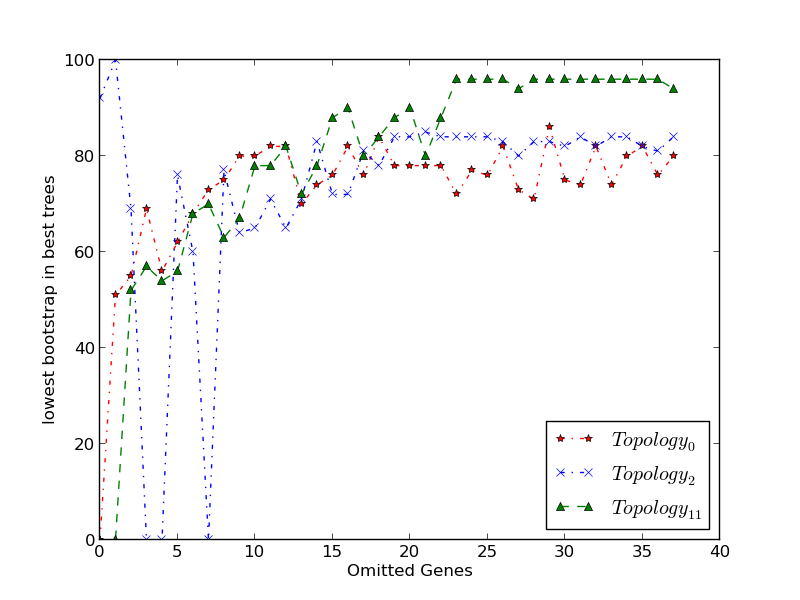}
    }
    \subfloat[\label{subfig:albs}]{%
      \includegraphics[width=0.5\textwidth]{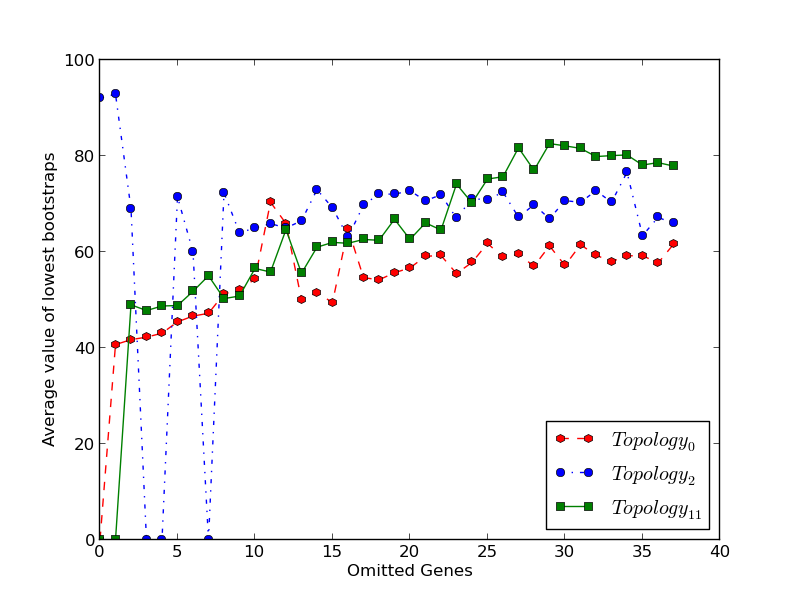}
    }
    \caption{Different comparisons of the topologies w.r.t the amount of removed genes: the number of disregarded genes in these figures is specified by $\frac{n}{3}$ where $n$ is the number of core genes. (a) Number of trees per topology, (b) number of trees whose lowest bootstrap is larger than or equal to 80, (c) lowest bootstrap in best trees, and (d) the average of lowest bootstraps.}
    \label{fig:statistics}
\end{figure}

As it can be noted, only 3 of the 43 obtained topologies contain trees whose lowest bootstrap is larger than 87, namely 0, 11, and 2. It is not so easy to make the decision, since all selected trees are very closed to each other with small differences. A new question needs to be answered: which genes are responsible for changing the tree from $\text{topology}_0$ to $\text{topology}_{11}$, or to $\text{topology}_2$? Deep investigations are needed in future work to answer this new question and to discover the set of genes in \emph{$group_A$, $group_B$, and $group_C$} that change one tree topology to another one (see Figure~\ref{fig:topologies}).    

The only notable difference between topologies 0 and 11 is the taxa position of \textit{Kalo\_septemlobus} and \textit{Meta\_delavayi}. In the same way, there is only one difference between topologies 0 and 11 with 2: grouping the same two taxa of \textit{Kalo\_septemlobus} and \textit{Meta\_delavayi}. Different comparisons on trees provided with selected topologies are summarized in Figure~\ref{fig:statistics}. 

\section{Conclusion}\label{con}
In this study, an many stages pipeline have been applied (namely: systematic mode, random mode, GA stage one, Lasso test mode, and GA stage two) for inferring trustworthy phylogenetic trees from various plant groups. We have verified that inferring a phylogenetic tree based on either the full set or some subsets of common core genes does not always lead to good support of the phylogenetic reconstruction. In both systematic and random stages, many trees have been generated based on omitting randomly some genes. When the desired score was not reached, a genetic algorithm has then been applied inside two specific stages using previously generated trees, to find new optimized solutions after realizing crossover and mutation operations. Furthermore, we applied a Lasso test for identifying and removing systematically blurring genes, discarding so those which have the worst impact on supports. 
We tested this pipeline on 322 different plant groups, where 63 of them are base families while the remaining ones are random trees, these latter playing the rule of skeletons when reconstructing the supertree. A case study regarding \emph{Apiales} order is analyzed and three ``best'' topologies stand out from the 43 obtained. Deep investigation will be needed in future work, in order to discover which genes change the topology, and to deeply investigate the sequences of the genes that blur the signal, to find the reasons of such effects.


\section*{Acknowledgement}
\textit{Computations have been performed on the supercomputer facilities of the M\'esocentre de calcul de Franche-Comt\'e.}

\bibliographystyle{unsrt}

\bibliography{biblio}

\end{document}